\documentclass[11pt,a4paper]{article}
\usepackage[hyperref]{emnlp2018}
\usepackage{times}
\usepackage{latexsym}

\usepackage{url}

\usepackage{amsmath}
\usepackage{graphicx}
\usepackage[colorinlistoftodos]{todonotes}
\usepackage{tabularx}
\usepackage{bbm}
\usepackage{times}
\usepackage{latexsym}
\usepackage{graphicx}
\usepackage{booktabs}
\usepackage{url}
\usepackage{comment}
\usepackage[english]{babel}
\usepackage[autostyle, english=american]{csquotes}
\MakeOuterQuote{"}
\usepackage[utf8x]{inputenc}

\aclfinalcopy 
\def\aclpaperid{66}

\title{SafeCity: Understanding Diverse Forms of Sexual Harassment \\ Personal Stories}

\author{Sweta Karlekar $\;\;\;\;\;\;\;\;$ Mohit Bansal \\
  UNC Chapel Hill \\
  {\tt \{swetakar, mbansal\}@cs.unc.edu}   }

\date{}

\begin{document}
\maketitle

\begin{abstract}
With the recent rise of \#MeToo, an increasing number of personal stories about sexual harassment and sexual abuse have been shared online. In order to push forward the fight against such harassment and abuse, we present the task of automatically categorizing and analyzing various forms of sexual harassment, based on stories shared on the online forum SafeCity. 
For the labels of groping, ogling, and commenting, our single-label CNN-RNN model achieves an accuracy of 86.5\%, and our multi-label model achieves a Hamming score of 82.5\%. 
Furthermore, we present analysis using LIME, first-derivative saliency heatmaps, activation clustering, and embedding visualization to interpret neural model predictions and demonstrate how this extracts features that can help automatically fill out incident reports, identify unsafe areas, avoid unsafe practices, and `pin the creeps'. 
\end{abstract}
\section{Introduction}

The hashtag \#MeToo\footnote{\url{https://metoomvmt.org}} has been prevalent on various social media platforms as a campaign centered around sharing stories of sexual harassment in an act of solidarity with other victims and spreading awareness of a widespread and endemic issue. 
With vast amounts of personal stories on the internet, it is important that we make scientific use of this data to push these movements forward and enable real-world change. Manually sorting and comprehending the information shared in these stories is an arduous task, and the power of natural language processing (NLP) can serve as the missing link between online activism and real change. 

We present several neural NLP models that allow us to automatically classify, aggregate, and analyze vast amounts of harassment data found on social media, becoming an effective tool for spreading awareness, increasing understanding, and allowing faster action. 
This large-scale automatic categorization, summarization, and analysis of personal abuse stories can help activist groups enlighten the public and advocate for social change in a timely manner.

We present single-label and multi-label classification of diverse forms of sexual harassment present in abuse stories shared online through the forum SafeCity, a crowd-sourcing platform for personal stories of sexual harassment and abuse.
Each story includes one or more tagged forms of sexual harassment, along with a description of the occurrence. For example, 
the description ``\emph{My college was nearby. This happened all the time. Guys passing comments, staring, trying to touch. Frustrating}'' is positive for three classes: commenting, ogling/staring, and touching/groping. 

 \begin{figure}[t]
 \centering
 \small
 \includegraphics[width=0.44\textwidth]{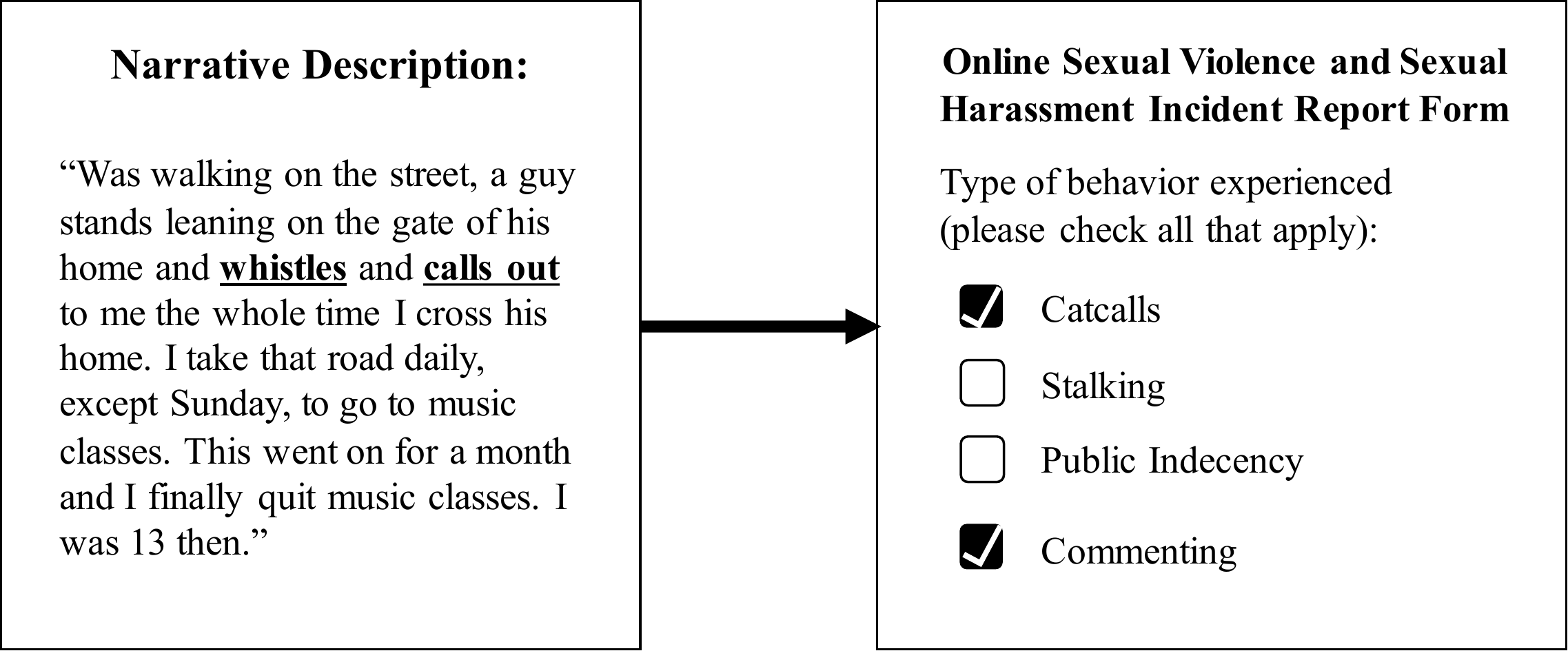}
 \vspace{-10pt}
\caption{\label{fig:task_descript} Task of sexual harassment story classification to help fill online incident reports. \vspace{-10pt}
}
 \end{figure}

We use CNN-RNN architectures (with character-level CNN embeddings and bidirectional RNNs) to classify the three forms of sexual harassment mentioned above using both single- and multi-label setups.
Our models achieve strong performances of 80-86\% on these setups.
This automatic classification of different forms of sexual harassment can help victims and authorities to partially automate and speed up the process of filling online sexual violence reporting forms (see Figure \ref{fig:task_descript}), which usually requires the victim to detail each form of sexual harassment that took place. The act of partially filling out the report (by our classifier) in itself makes it more likely for the victim to file a report. A study by the Bureau of Justice found that victims who report sexual assault are more likely to seek medical treatment for injuries, which also allows for more immediate prosecution and a better chance of finding DNA evidence to convict the offender \cite{rennison2002rape}. Further, it can also be used to fulfill the need to automatically categorize and summarize large numbers of online testimonials describing or reporting sexual harassment.

Next, in order to further utilize these stories as an important tool for harassment understanding and to help prevent similar situations from happening to others, we present interpretability analysis of our neural classification results in the forms of LIME analysis, first-derivative saliency heatmaps, activation clustering, and t-SNE embedding visualization. We show how these analysis techniques hold promise as avenues for future work and can potentially provide insightful clues towards building (1) a tool to analyze the most common circumstances around each distinct form of harassment to provide more detailed and accurate safety advice, (2) a map of unsafe areas to help others avoid dangerous spaces, and 3) an unofficial sex offender registry that marks frequently-mentioned offenders to warn potential victims. This paper seeks to provide an avenue to utilize the millions of stories shared on social media describing instances of sexual harassment, including \#MeToo, \#WhyILeft, and \#YesAllWomen. With this task and analysis, we hope that these stories can be used to prevent future sexual harassment.

\section{Related Work}
\vspace{-8pt}
Analyzing personal sexual harassment stories from online social forums is fairly unexplored, to the best of our knowledge. However, recent works in a similar vein include detecting the presence of domestic abuse stories on social media sites~\cite{schrading2015analysis,schrading2015analyzing,schrading2015whyistayed}. In more distantly related work, NLP has been used for various socially-driven tasks, such as detecting the presence of cyberbullying or incivility~\cite{TUD-CS-2018-0023,founta2018unified,chen2012detecting,zhao2016automatic,agrawal2018deep,van2018automatic}, and detecting and providing aid for signs of depression or suicidal thoughts~\cite{pestian2010suicide,yazdavar2017semi,stepanov2017depression,fitzpatrick2017delivering}.

\section{Classification Models}
\vspace{-4pt}
\label{sect: Classification Models}

For our single-label binary classification task, the two output classes can be [commenting, non-commenting], [ogling, non-ogling], or [groping, non-groping]. For our multi-label scenario, there are a total of 8 combinations (true or false for three types of sexual harassment), including a label for none of the three classes present in the description.

\noindent\textbf{CNN}:
For each input description, an embedding and convolutional layer are applied. This is followed by a max-pooling layer~\cite{collobert2011natural}. Filters of varying window sizes are applied to each window of word vectors, the result of which is then passed through a softmax layer to produce probabilities over the output classes. 

 \begin{figure}[t]
 \centering
 \includegraphics[width=0.32\textwidth]{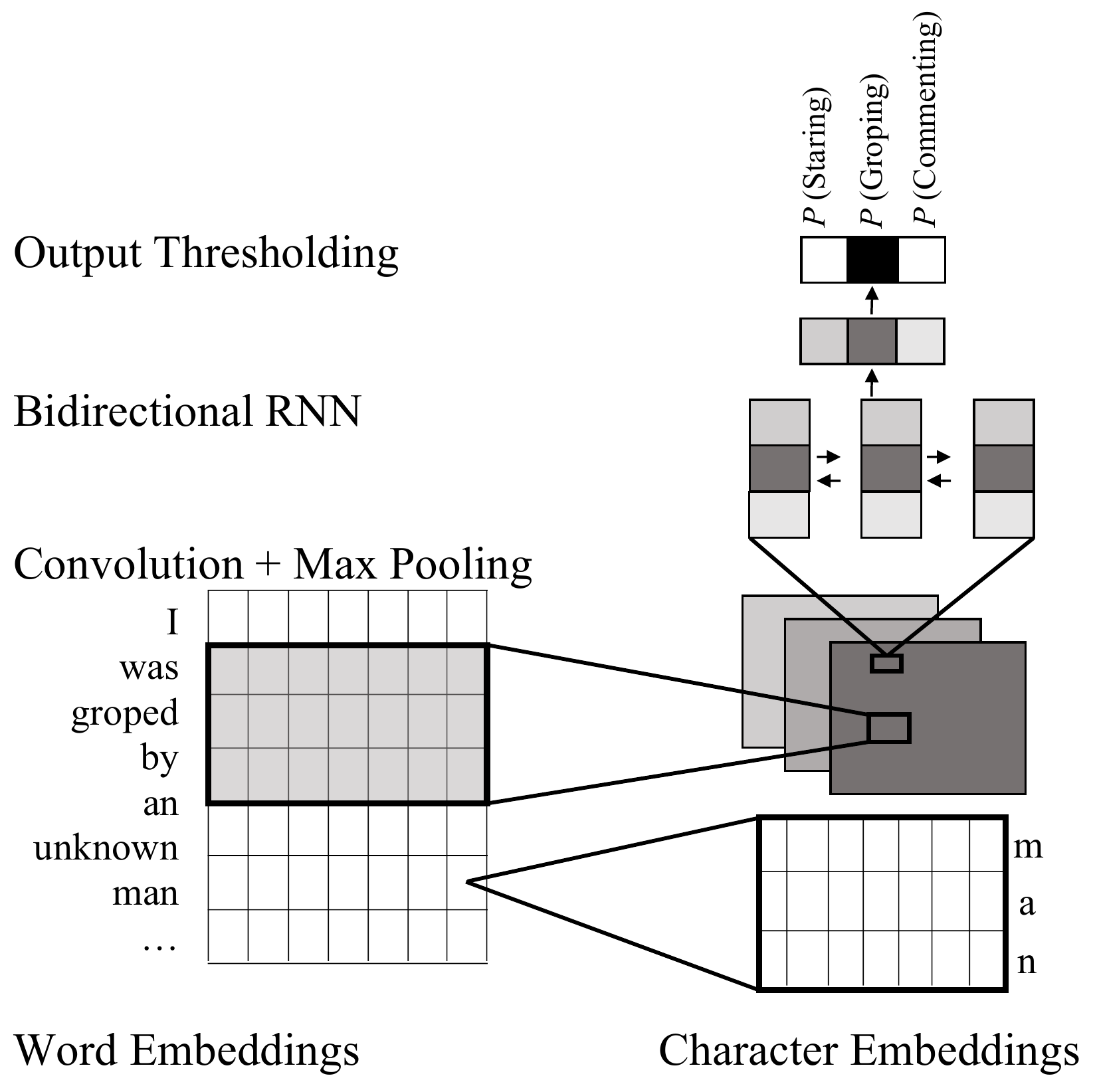}
\vspace{-10pt}
\caption{\label{fig:cnnrnn} Multi-label CNN-RNN model with CNN-based character embeddings and bidirectional RNNs.
}
\vspace{-12pt}
\end{figure}

\noindent\textbf{LSTM-RNN}:
As CNNs are not designed to capture sequential relationships~\cite{pascanu2013construct}, we adopted an RNN model that consisted of word vectors fed into LSTM layer, the final state of which was fed into a fully-connected layer. The result is passed through a softmax layer to output the probability over all output classes. 

\noindent\textbf{CNN-RNN}:
As both models have strengths and weaknesses, we experimented with a hybrid architecture in which our LSTM-RNN model after the embedding layer is laid on top of our CNN model before the max-pooling (related to~\newcite{zhou2015c}). For single-label models, the final fully-connected layer is fed into a softmax to give final output probabilities.

\noindent\textbf{Multi-Label Classification}
We also present multi-label classification~\cite{boutell2004learning,tsoumakas2006multi,katakis2008multilabel}, 
which allows for models to predict multiple categories simultaneously for the same input. We further utilized CNN-based character embeddings in addition to word embeddings, and also employed bidirectional RNNs (see Figure \ref{fig:cnnrnn}). The outputs of the final fully-connected layer (\textit{F}) are fed into a sigmoid function. The classification for each category (C) are seen as positive (1) if the output is above threshold \textit{t} and negative (0) if the output is below threshold \textit{t}, a hyperparameter, giving the equation: $C = \mathbbm{1}(\sigma(F) \geq t)$.

\section{Experimental Setup}
\vspace{-7pt}
\subsection{Dataset} 
SafeCity\footnote{\url{http://safecity.in}} is, to the best of our knowledge, the largest publicly-available online forum for reporting sexual harassment. Its motto is "pin the creeps". 
Victims of sexual harassment share personal stories, with the objective of spreading awareness of ongoing sexual harassment and showcasing location-based trends.
The language styles of SafeCity forums are very diverse, and therefore can potentially be used for a variety of test cases, such as emails or tweets.

Each of the 9,892 stories includes a description of the incident, the location, and tagged forms of harassment, with all identifying information removed. 
SafeCity has explicitly given us permission to use this data.
The dataset\footnote{We release our dataset splits at \url{https://github.com/swkarlekar/safecity}. Please follow SafeCity guidelines for usage.} contains descriptions of text submitted by forum users, along with tags of 13 forms of sexual harassment. 
We chose the top three most dense categories---groping/touching, staring/ogling, and commenting---to use as our dataset, as the others were more sparse. Each description may fall into none, some, or all of the categories. 

\subsection{Evaluation}
The single-label models were evaluated using accuracy. The multi-label models were evaluated using exact match ratio and Hamming score (calculated as the complement of Hamming loss). Hamming loss was used as detailed by~\newcite{tsoumakas2006multi}.
Hamming loss (\textit{y}) is equal to 1 over $|D|$ (number of multi-label samples), multiplied by the sum of the symmetric differences between the predictions (\textit{Z}) and the true labels (\textit{Y}), divided by the number of labels (\textit{L}), giving $y = \frac{1}{|D|}\sum\limits_{i=1}^{|D|}\frac{|Y_i\Delta Z_i|}{|L|}$. 

\subsection{Training Details}
All models have vocabulary size of $10,000$, and use AdamOptimizer~\cite{kingma2014adam} with a learning rate of $1e^{-4}$. All gradient norms are clipped to $2.0$~\cite{pascanu2013difficulty,graves2013generating}.
For each model, the hyperparameters are tuned using the development set.

\vspace{-5pt}
\paragraph{CNN}
We use a $2$-D CNN.
Filter sizes of [$3$, $4$, $5$] are used with $128$ filters per filter size. 
Batch size is set to $128$, and a dropout~\cite{srivastava2014dropout} of $0.80$ is applied.

\vspace{-5pt}
\paragraph{LSTM}
Our LSTM has $2$ layers with $60$ hidden units. 
Batch size is $64$ with a dropout of $0.75$.

\vspace{-5pt}
\paragraph{CNN-LSTM}
Our CNN-LSTM model consists of an LSTM on top of a CNN. The CNN has 100 filters per filter size of [$3$, $4$, $5$]. Embedding dimensions of $300$ are used. An LSTM with $300$ hidden units is used. For the character level embeddings, we use an additional CNN with 100 filters per filter size of [$3$, $4$, $5$]. Bidirectional RNNs of $300$ units are used.

\begin{table}[t]
\centering
\begin{small}
\begin{tabular}{r|c|c|c}
Model & Commenting & Ogling & Groping \\\hline
Linear SVM & 42.2 & 35.0 & 55.8 \\
Gaussian NB & 46.8 & 74.7 & 66.0 \\
Logistic Reg. & 61.4 & 78.0 & 69.1 \\
SVM & 65.5 & 79.0 & 70.3 \\
CNN & 80.9 & 82.2 & 86.0 \\
RNN & 81.0 & 82.2 & 86.2 \\
CNN-RNN & \textbf{81.6} & \textbf{84.1} & \textbf{86.5} \\
\end{tabular}
\end{small}
\vspace{-7pt}
\caption{\label{tab:single-label-results} Single-label classification (accuracy) results. \vspace{-2pt}
}
\end{table}

\vspace{-7pt}
\section{Results}
\vspace{-7pt}
See Table \ref{tab:single-label-results} for single-label results on the selected harassment categories, where CNN-RNN was the best performing model compared to several non-neural and neural baselines. See Table \ref{tab:multi-label-results} for multi-label classification results, where the Hamming score for the multi-label CNN-RNN model is 82.5\%, showing potential for real-world use as well as substantial future research scope.

\begin{table}[t]
\centering
\begin{small}
\begin{tabular}{r|c|c}
Model & Exact Match & Hamming \\\hline
Random Forest & 35.0 & 70.2 \\
CNN & 53.7 & 80.2 \\
RNN & 57.1 & 81.5 \\
CNN-RNN & 59.2 & 82.3 \\
CNN-RNN (bidirec + char) & \textbf{62.0} & \textbf{82.5}
\end{tabular}
\end{small}
\vspace{-13pt}
\caption{\label{tab:multi-label-results} Multi-label classification results. \vspace{-10pt}
}
\end{table}

\begin{figure*}[ht]
\minipage{0.29\textwidth}
\includegraphics[width=\textwidth]{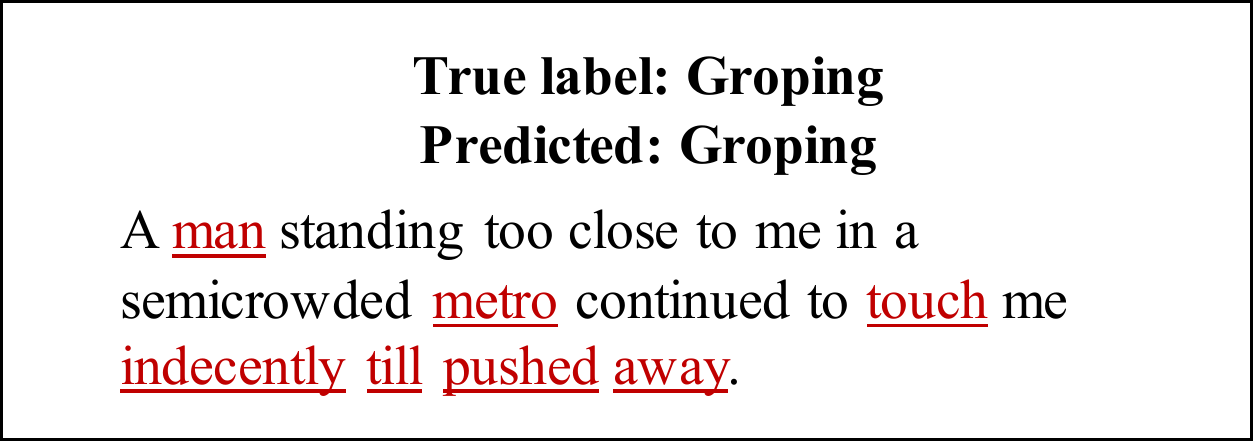}
\endminipage\hfill
\minipage{0.29\textwidth}
 \includegraphics[width=\textwidth]{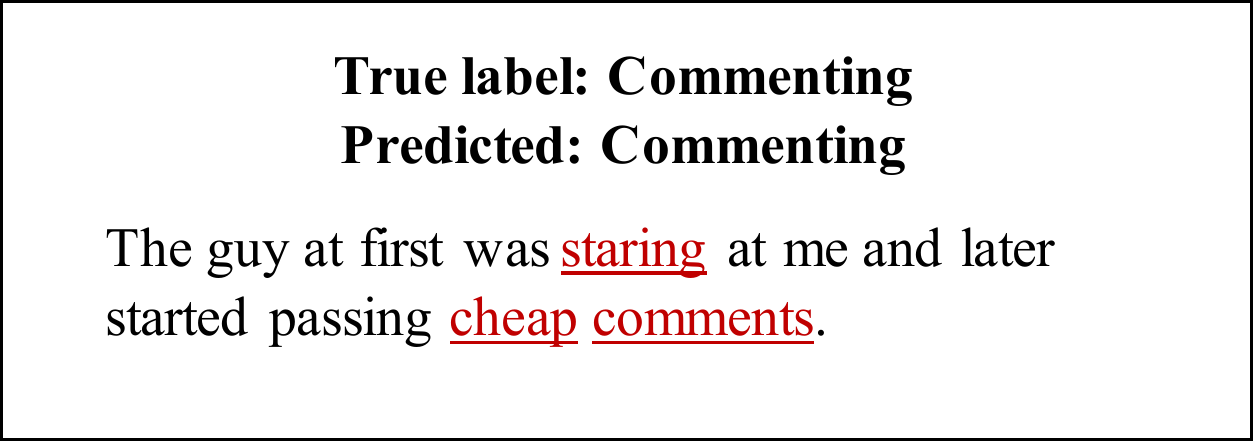}
\endminipage\hfill
\minipage{0.29\textwidth}%
 \includegraphics[width=\textwidth]{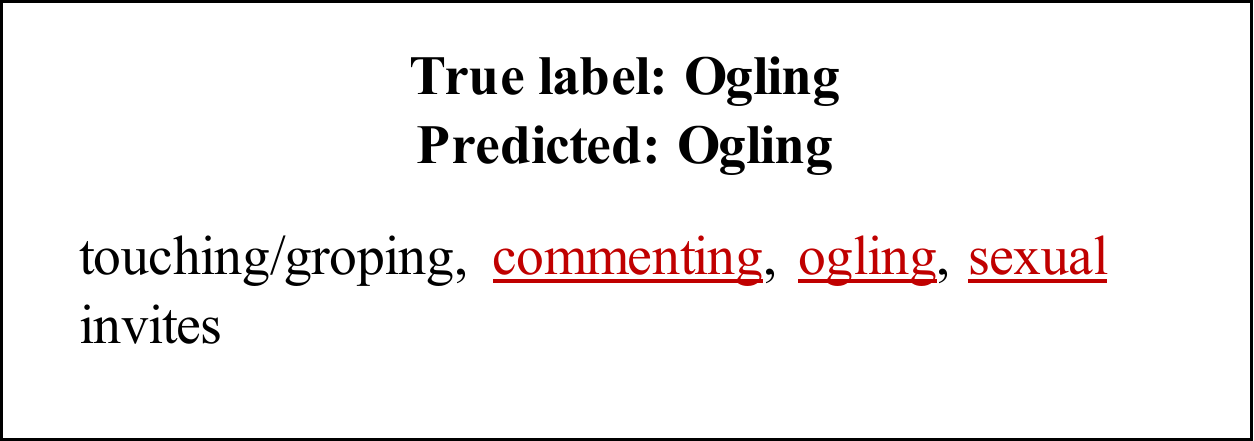}
\endminipage
\vspace{-7pt}
\caption{
LIME analysis. 
\textbf{Left }:
Correctly-classified example of groping.  
\textbf{Middle}: Correctly-classified example of commenting. 
\textbf{Right}: Correctly-classified example of ogling/staring. 
\vspace{-10pt}}
\label{fig:examples}
\end{figure*}

\begin{figure*}[ht]
\minipage{0.32\textwidth}
\includegraphics[width=\textwidth]{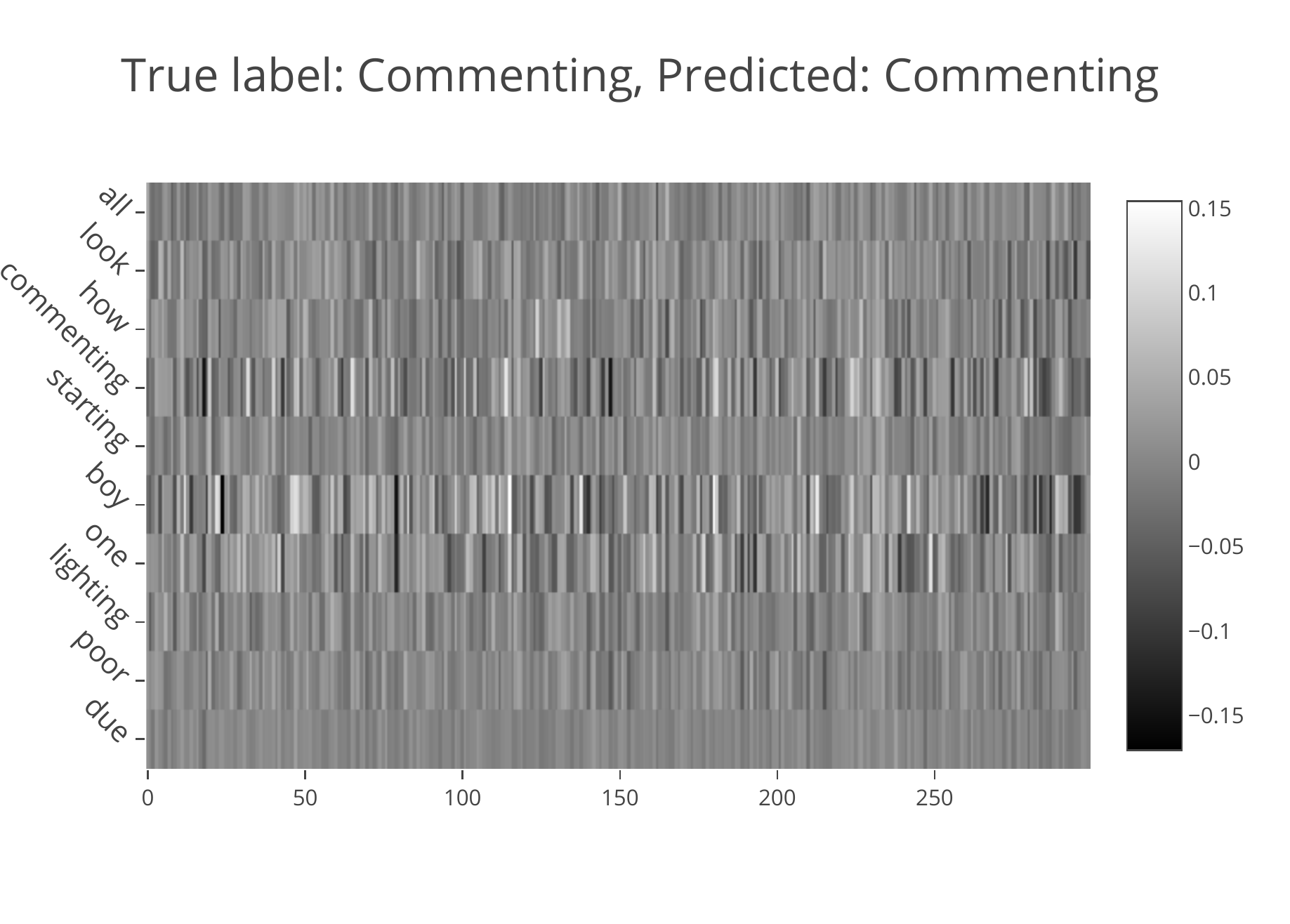}
\endminipage\hfill
\minipage{0.32\textwidth}
 \includegraphics[width=\textwidth]{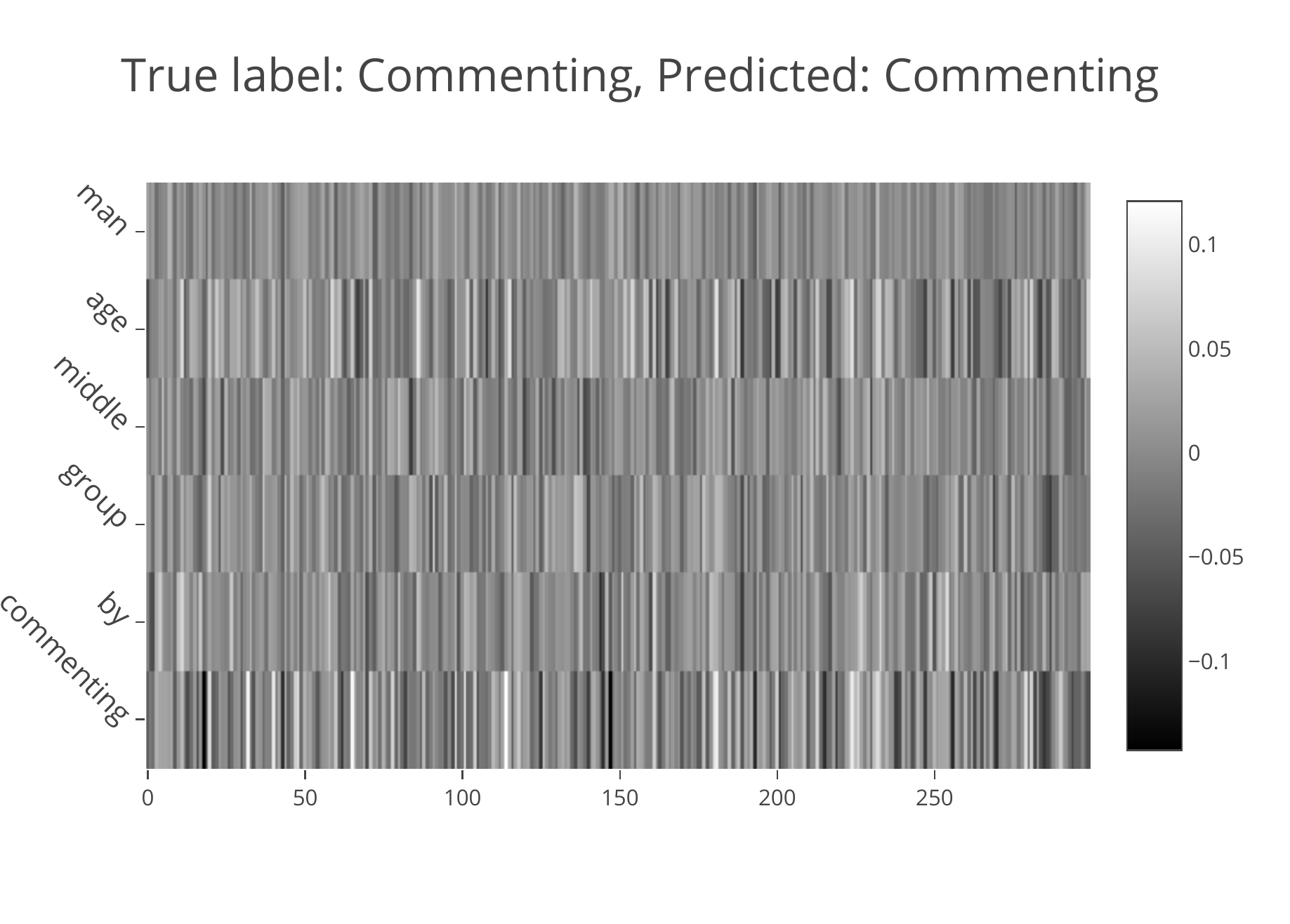}
\endminipage\hfill
\minipage{0.32\textwidth}%
 \includegraphics[width=\textwidth]{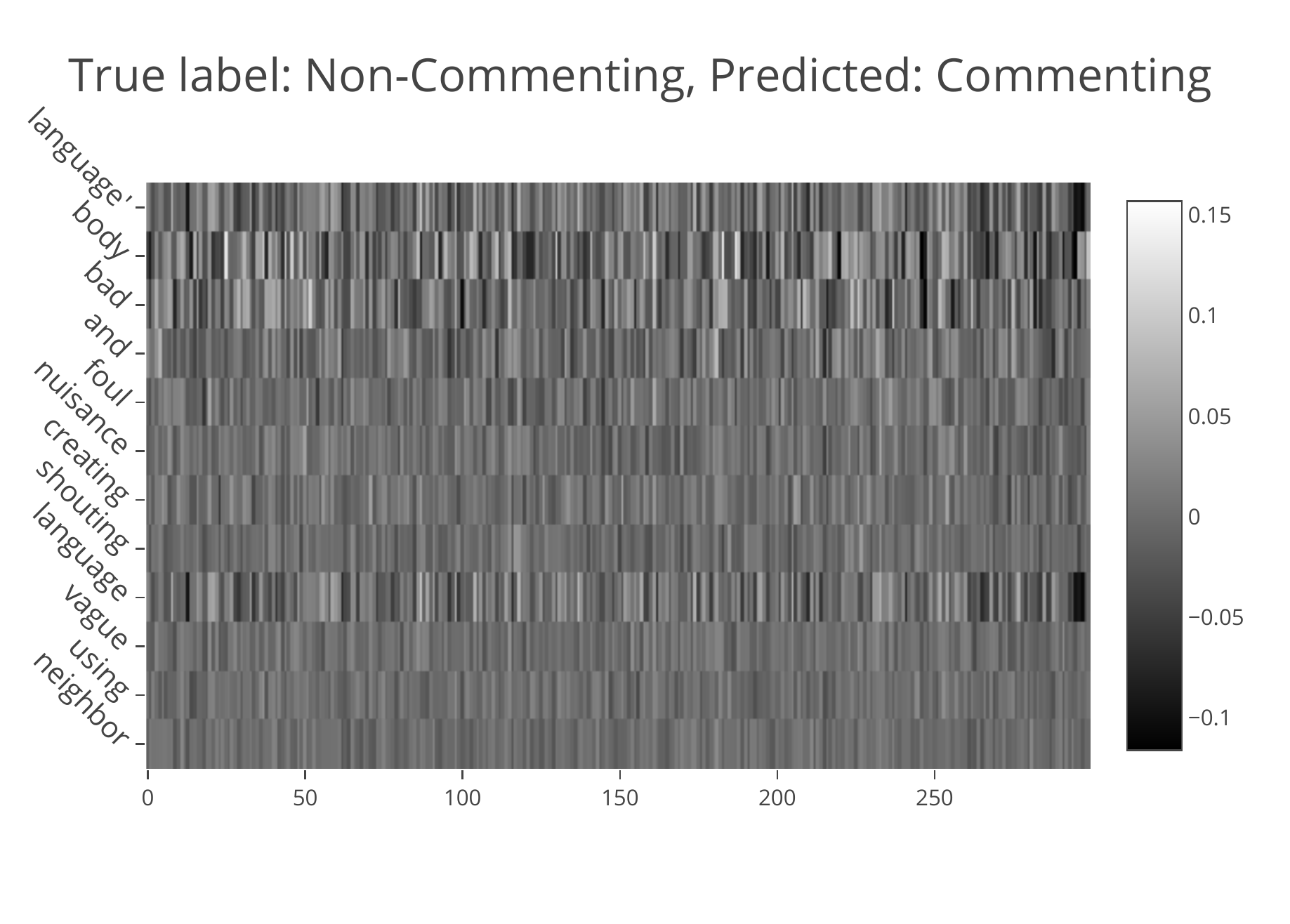}
\endminipage
\vspace{-18pt}
\caption{
First derivative saliency heatmaps. 
\textbf{Left}:
Correctly-classified example of commenting.
\textbf{Middle}: Correctly-classified example of commenting.
\textbf{Right}: Incorrectly-classified example of commenting. 
\vspace{-12pt}
}
\label{fig:heatmaps}
\end{figure*}

\begin{table}[t]
\centering
\label{tab: word_embeddings}
\begin{small}
\begin{tabular}{|l|p{4.4cm}|}
\hline
Harassment Type & Observed t-SNE Cluster\\\hline
Groping & abusively, encounter, talk, underage, surrounded, embarrassed\\\hline
Ogling & boobs, leering, disturbing, gestures, voyeur, visually, gestures  \\\hline
Commenting & shameful, vulgar, inappropriately, indecent, invites, stalked, strong  \\\hline
\end{tabular}
\end{small}
\vspace{-7pt}
\caption{\label{tab:word_embeddings} Relevant word clusters found via t-SNE word embedding clustering. \vspace{-10pt}
}
\end{table}

\vspace{-6pt}
\section{Analysis}
\vspace{-6pt}
We provide various visualization techniques to analyze our models. Each of these techniques employs a different approach and offers new information or supports previous findings.

\vspace{-2pt}
\subsection{Word Embedding Visualization}
\vspace{-2pt}
We selected seed words that corresponded to class labels and found the nearest neighbors of each seed word's vector by reducing the dimensionality of the word embeddings using t-SNE (see Table \ref{tab:word_embeddings})~\cite{maaten2008visualizing}. This form of visualization not only ensures that our model has learned appropriate word embeddings, but also demonstrates that each form of sexual harassment has a unique and distinct context. Furthermore, this shows that our model learns related words and concepts for each type of harassment.

\vspace{-2pt}
\subsection{LIME Analysis}
\vspace{-2pt}
LIME analysis~\cite{ribeiro2016should}, or Local Interpretable Model-Agnostic Explanation, interprets the local reasoning of a model around an instance.
Results of LIME ($\xi$) are found by taking the minimum of $\mathcal{L}$, which is the measure of how unfaithful the interpretable model (\textit{g}) is to approximating the probability that an input (\textit{x}) belongs to a certain class (\textit{f}) in the locally defined area ($\pi_x$) summed with complexity measures $\Omega$, giving $\xi(x) = argmin~ \mathcal{L}(f, g, \pi_x) + \Omega(g)$.
In Figure \ref{fig:examples} (left), the words "touch", "man", and the collective words "indecently till pushed away" are the most important to the local classification of "groping". Furthermore, the word "metro" has importance in the classification, suggesting that this may be a frequent location in which groping takes place. In Figure \ref{fig:examples} (middle), the words with the most importance are "comments" and "staring", indicating that ogling may coincide with commenting very frequently. In Figure \ref{fig:examples} (right), the words "ogling", "sexual", and "commenting" had the most importance, which further supports the notion that ogling and commenting often occur together. As verified by the data, ogling and commenting together is more common than ogling alone. 

\vspace{-5pt}
\subsection{First Derivative Saliency}
\vspace{-5pt}
Saliency heatmaps~\cite{simonyan2013deep,li2015visualizing} 
illustrate which words of an input have the biggest impact on the final classification by taking the gradient of the final scores outputted by the neural network (\textit{S}) with respect to the embedding (\textit{E}), given the true label (\textit{L}), giving $\frac{\partial S_L(E)}{\partial E}$. 
While LIME analysis and first derivative saliency are both used to find word-level contributions, first derivative saliency is model-dependent and gives reasoning behind classification based on the whole model, in contrast to the locally-faithful, model-agnostic LIME analysis technique.

In Figure \ref{fig:heatmaps} (left), the word "commenting" and the words "one boy" have the most influence on the classification. The influence of the word "lighting" indicates poor lighting is often present in situations where sexual harassment takes place.
In Figure \ref{fig:heatmaps} (middle), the classification of "commenting" was most influenced by the word "commenting", followed by the word "age". This suggests the possibility of using descriptors of offenders as a classification tool. 
Figure \ref{fig:heatmaps} (right) is an incorrectly classified example. We see that the word "body", followed by "language", had the most influence on the classification of this example as "commenting". Our model identifies synonyms and hyponyms like the word "language" in relation to the category of commenting. However, the true label was "non-commenting", as the word was not used in a context of sexual language, but rather as "vague language" and "body language".

\vspace{-4pt}
\subsection{Activation Clustering}
\vspace{-4pt}
Activation clustering \cite{girshick2014rich,aubakirova2016interpreting} accesses the activation values of all \textit{n} neurons and treats the activation values per input as coordinates in \textit{n}-dimensional space. K-means clustering was performed to group activation clusters and find common themes in these reports. Activation clustering is distinct from both LIME analysis and first derivative saliency in that it finds patterns and clusterings at a description-level.

\noindent\textbf{Circumstances of Harassment}:
One of the clusters was classified as "ogling": \{\textit{`a group of boys was standing near us and were making weird expressions and as we moved away they started following'; `a group of guys lurking around the theater...'}\}. 
Another cluster was classified as "commenting": \{\textit{'a group of men were standing who commented on every girl who passed by the', 'a group of boys were standing there... as we started moving one of them commented on us'}\} 
Both of these clusters contained examples describing circumstances of the harassment, following the pattern of "a group of boys/men were standing/lurking and..." 
It can be inferred that certain forms of sexual harassment are more likely to happen with large groups of men. 
Activation clustering can identify the circumstances of harassment, helping potential victims to be better prepared. 

\noindent\textbf{Location and Time of Harassment}:
Some clusters contain examples that point to specific locations of harassment, e.g., a groping cluster: \{\textit{`i was in the bus and there was this man who purposely fell on me and touched me inappropriately'; `while traveling in a crowded bus most of the time men try to grind their intimate part over my body'; `i was in the bus when a man standing tried to put his d**k on my hand'}\}.
Specific locations can also be found: \{\textit{`the gurgaon sohna road is very unsafe at night especially if you are alone with no street lights'; `kurla station really gets scary at night once i was trying to get a train from kurla station around 10'; `mathura highway , not enough lights on the way during nights so is not safe for a individual to journey'}\}.
Notice that the second cluster examples also contain the word "night". With data that contains more specific locations or times of day, activation clusters can
serve as an automatic way to map out unsafe areas based on location and time of day.

\noindent\textbf{Identifying Offenders}:
Examples from another groping cluster include: \{\textit{`...her step father abused her physically for a year'; `one of the girl of about 6 years got raped by her own father'; `it happened at my house my brother harassed me and also misbehaved with me one night its been six months'}\}.
This shows that clusters can point to common relationships or titles for offenders. This phenomenon can be presumed to happen with names of offenders as well. 
If many reports have been filed around this offender, clusters will form around his/her name. 
Instead of a case of "he said, she said", activation clustering provides an avenue towards "he said, they said", as clusters form when multiple reports have been filed around the same name.

The main purpose of our visualization techniques is to explain what the black-box deep learning models are learning, such as locations, offenders, or times of day. With more detailed data in the future, we may be able to uncover more nuanced circumstances behind harassment.

\vspace{-6pt}
\section{Conclusion}
\vspace{-6pt}
We presented the novel task of identifying various forms of sexual harassment in personal stories. Our accurate multi-label classification models illustrate the plausibility of automatically filling out incident reports.
Using visualization techniques, we found circumstances surrounding forms of harassment and the possibility of automatically identifying safe areas and repeat offenders. 
In future work, we hope to experiment with the transferability of our model to other datasets to encompass the diverse mediums through which these personal stories are shared. Honoring the courage that these victims demonstrated in sharing their stories online, we use these descriptions not only to help summarize online testimonials and provide more detailed safety advice, but also to help others report similar occurrences to hopefully prevent future sexual harassment from occurring.

\section*{Acknowledgments}
We thank the SafeCity moderators for their assistance with the data download, and the anonymous reviewers for their helpful comments. This work was supported by a
Google Faculty Research Award, a Bloomberg Data Science Research Grant, an IBM Faculty
Award, and NVidia GPU awards.

\bibliography{acl2018}

\begin{thebibliography}{30}
\expandafter\ifx\csname natexlab\endcsname\relax\def\natexlab#1{#1}\fi

\bibitem[{Agrawal and Awekar(2018)}]{agrawal2018deep}
Sweta Agrawal and Amit Awekar. 2018.
\newblock Deep learning for detecting cyberbullying across multiple social
  media platforms.
\newblock \emph{European Conference on Information Retrieval}.

\bibitem[{Aubakirova and Bansal(2016)}]{aubakirova2016interpreting}
Malika Aubakirova and Mohit Bansal. 2016.
\newblock Interpreting neural networks to improve politeness comprehension.
\newblock \emph{Proceedings of EMNLP}.

\bibitem[{Boutell et~al.(2004)Boutell, Luo, Shen, and
  Brown}]{boutell2004learning}
Matthew~R Boutell, Jiebo Luo, Xipeng Shen, and Christopher~M Brown. 2004.
\newblock Learning multi-label scene classification.
\newblock \emph{Pattern Recognition}, 37(9):1757--1771.

\bibitem[{Chen et~al.(2012)Chen, Zhou, Zhu, and Xu}]{chen2012detecting}
Ying Chen, Yilu Zhou, Sencun Zhu, and Heng Xu. 2012.
\newblock Detecting offensive language in social media to protect adolescent
  online safety.
\newblock In \emph{Privacy, Security, Risk and Trust (PASSAT), 2012
  International Conference on and 2012 International Conference on Social
  Computing (SocialCom)}, pages 71--80. IEEE.

\bibitem[{Collobert et~al.(2011)Collobert, Weston, Bottou, Karlen, Kavukcuoglu,
  and Kuksa}]{collobert2011natural}
Ronan Collobert, Jason Weston, L{\'e}on Bottou, Michael Karlen, Koray
  Kavukcuoglu, and Pavel Kuksa. 2011.
\newblock Natural language processing (almost) from scratch.
\newblock \emph{Journal of Machine Learning Research}, 12(Aug):2493--2537.

\bibitem[{Fitzpatrick et~al.(2017)Fitzpatrick, Darcy, and
  Vierhile}]{fitzpatrick2017delivering}
Kathleen~Kara Fitzpatrick, Alison Darcy, and Molly Vierhile. 2017.
\newblock Delivering cognitive behavior therapy to young adults with symptoms
  of depression and anxiety using a fully automated conversational agent
  (woebot): a randomized controlled trial.
\newblock \emph{JMIR Mental Health}, 4(2).

\bibitem[{Founta et~al.(2018)Founta, Chatzakou, Kourtellis, Blackburn, Vakali,
  and Leontiadis}]{founta2018unified}
Antigoni-Maria Founta, Despoina Chatzakou, Nicolas Kourtellis, Jeremy
  Blackburn, Athena Vakali, and Ilias Leontiadis. 2018.
\newblock A unified deep learning architecture for abuse detection.
\newblock \emph{CoRR}.

\bibitem[{Girshick et~al.(2014)Girshick, Donahue, Darrell, and
  Malik}]{girshick2014rich}
Ross Girshick, Jeff Donahue, Trevor Darrell, and Jitendra Malik. 2014.
\newblock Rich feature hierarchies for accurate object detection and semantic
  segmentation.
\newblock In \emph{Proceedings of the IEEE Conference on Computer Vision and
  Pattern Recognition}, pages 580--587.

\bibitem[{Graves(2013)}]{graves2013generating}
Alex Graves. 2013.
\newblock Generating sequences with recurrent neural networks.
\newblock \emph{arXiv preprint arXiv:1308.0850}.

\bibitem[{Katakis et~al.(2008)Katakis, Tsoumakas, and
  Vlahavas}]{katakis2008multilabel}
Ioannis Katakis, Grigorios Tsoumakas, and Ioannis Vlahavas. 2008.
\newblock Multilabel text classification for automated tag suggestion.
\newblock In \emph{Proceedings of the ECML/PKDD}, volume~18.

\bibitem[{Kingma and Ba(2015)}]{kingma2014adam}
Diederik Kingma and Jimmy Ba. 2015.
\newblock Adam: A method for stochastic optimization.
\newblock \emph{3rd International Conference for Learning Representations}.

\bibitem[{Li et~al.(2016)Li, Chen, Hovy, and Jurafsky}]{li2015visualizing}
Jiwei Li, Xinlei Chen, Eduard Hovy, and Dan Jurafsky. 2016.
\newblock Visualizing and understanding neural models in nlp.
\newblock \emph{Proceedings of NAACL}.

\bibitem[{Maaten and Hinton(2008)}]{maaten2008visualizing}
Laurens van~der Maaten and Geoffrey Hinton. 2008.
\newblock Visualizing data using t-sne.
\newblock \emph{Journal of Machine Learning Research}, 9(Nov):2579--2605.

\bibitem[{Pascanu et~al.(2014)Pascanu, Gulcehre, Cho, and
  Bengio}]{pascanu2013construct}
Razvan Pascanu, Caglar Gulcehre, Kyunghyun Cho, and Yoshua Bengio. 2014.
\newblock How to construct deep recurrent neural networks.
\newblock \emph{Proceedings of ICLR}.

\bibitem[{Pascanu et~al.(2013)Pascanu, Mikolov, and
  Bengio}]{pascanu2013difficulty}
Razvan Pascanu, Tomas Mikolov, and Yoshua Bengio. 2013.
\newblock On the difficulty of training recurrent neural networks.
\newblock In \emph{International Conference on Machine Learning}, pages
  1310--1318.

\bibitem[{Pestian et~al.(2010)Pestian, Nasrallah, Matykiewicz, Bennett, and
  Leenaars}]{pestian2010suicide}
John Pestian, Henry Nasrallah, Pawel Matykiewicz, Aurora Bennett, and Antoon
  Leenaars. 2010.
\newblock Suicide note classification using natural language processing: A
  content analysis.
\newblock \emph{Biomedical Informatics Insights}, 3:BII--S4706.

\bibitem[{Rennison(2002)}]{rennison2002rape}
Callie~Marie Rennison. 2002.
\newblock \emph{Rape and sexual assault: Reporting to police and medical
  attention, 1992-2000}.
\newblock US Department of Justice, Office of Justice Programs Washington, DC.

\bibitem[{Ribeiro et~al.(2016)Ribeiro, Singh, and Guestrin}]{ribeiro2016should}
Marco~Tulio Ribeiro, Sameer Singh, and Carlos Guestrin. 2016.
\newblock Why should i trust you?: Explaining the predictions of any
  classifier.
\newblock In \emph{Proceedings of the 22nd ACM SIGKDD International Conference
  on Knowledge Discovery and Data Mining}, pages 1135--1144. ACM.

\bibitem[{Schrading(2015)}]{schrading2015analyzing}
J~Nicolas Schrading. 2015.
\newblock \emph{Analyzing domestic abuse using natural language processing on
  social media data}.
\newblock Rochester Institute of Technology.

\bibitem[{Schrading et~al.(2015{\natexlab{a}})Schrading, Alm, Ptucha, and
  Homan}]{schrading2015analysis}
Nicolas Schrading, Cecilia~Ovesdotter Alm, Ray Ptucha, and Christopher Homan.
  2015{\natexlab{a}}.
\newblock An analysis of domestic abuse discourse on reddit.
\newblock In \emph{Proceedings of the 2015 Conference on Empirical Methods in
  Natural Language Processing}, pages 2577--2583.

\bibitem[{Schrading et~al.(2015{\natexlab{b}})Schrading, Alm, Ptucha, and
  Homan}]{schrading2015whyistayed}
Nicolas Schrading, Cecilia~Ovesdotter Alm, Raymond Ptucha, and Christopher
  Homan. 2015{\natexlab{b}}.
\newblock \# whyistayed,\# whyileft: Microblogging to make sense of domestic
  abuse.
\newblock In \emph{Proceedings of the 2015 Conference of the North American
  Chapter of the Association for Computational Linguistics: Human Language
  Technologies}, pages 1281--1286.

\bibitem[{Simonyan et~al.(2014)Simonyan, Vedaldi, and
  Zisserman}]{simonyan2013deep}
Karen Simonyan, Andrea Vedaldi, and Andrew Zisserman. 2014.
\newblock Deep inside convolutional networks: Visualising image classification
  models and saliency maps.
\newblock \emph{ICLR Workshop}.

\bibitem[{Srivastava et~al.(2014)Srivastava, Hinton, Krizhevsky, Sutskever, and
  Salakhutdinov}]{srivastava2014dropout}
Nitish Srivastava, Geoffrey~E Hinton, Alex Krizhevsky, Ilya Sutskever, and
  Ruslan Salakhutdinov. 2014.
\newblock Dropout: a simple way to prevent neural networks from overfitting.
\newblock \emph{Journal of Machine Learning Research}, 15(1):1929--1958.

\bibitem[{Stepanov et~al.(2017)Stepanov, Lathuiliere, Chowdhury, Ghosh, Vieriu,
  Sebe, and Riccardi}]{stepanov2017depression}
Evgeny Stepanov, Stephane Lathuiliere, Shammur~Absar Chowdhury, Arindam Ghosh,
  Radu-Laurentiu Vieriu, Nicu Sebe, and Giuseppe Riccardi. 2017.
\newblock Depression severity estimation from multiple modalities.
\newblock \emph{arXiv preprint arXiv:1711.06095}.

\bibitem[{Tsoumakas and Katakis(2006)}]{tsoumakas2006multi}
Grigorios Tsoumakas and Ioannis Katakis. 2006.
\newblock Multi-label classification: An overview.
\newblock \emph{International Journal of Data Warehousing and Mining}, 3(3).

\bibitem[{Van~Hee et~al.(2018)Van~Hee, Jacobs, Emmery, Desmet, Lefever,
  Verhoeven, De~Pauw, Daelemans, and Hoste}]{van2018automatic}
Cynthia Van~Hee, Gilles Jacobs, Chris Emmery, Bart Desmet, Els Lefever, Ben
  Verhoeven, Guy De~Pauw, Walter Daelemans, and V{\'e}ronique Hoste. 2018.
\newblock Automatic detection of cyberbullying in social media text.
\newblock \emph{arXiv preprint arXiv:1801.05617}.

\bibitem[{Yazdavar et~al.(2017)Yazdavar, Al-Olimat, Ebrahimi, Bajaj, Banerjee,
  Thirunarayan, Pathak, and Sheth}]{yazdavar2017semi}
Amir~Hossein Yazdavar, Hussein~S Al-Olimat, Monireh Ebrahimi, Goonmeet Bajaj,
  Tanvi Banerjee, Krishnaprasad Thirunarayan, Jyotishman Pathak, and Amit
  Sheth. 2017.
\newblock Semi-supervised approach to monitoring clinical depressive symptoms
  in social media.
\newblock In \emph{Proceedings of the 2017 IEEE/ACM International Conference on
  Advances in Social Networks Analysis and Mining 2017}, pages 1191--1198. ACM.

\bibitem[{Zhao et~al.(2016)Zhao, Zhou, and Mao}]{zhao2016automatic}
Rui Zhao, Anna Zhou, and Kezhi Mao. 2016.
\newblock Automatic detection of cyberbullying on social networks based on
  bullying features.
\newblock In \emph{Proceedings of the 17th International Conference on
  Distributed Computing and Networking}, page~43. ACM.

\bibitem[{Zhou et~al.(2015)Zhou, Sun, Liu, and Lau}]{zhou2015c}
Chunting Zhou, Chonglin Sun, Zhiyuan Liu, and Francis Lau. 2015.
\newblock A c-lstm neural network for text classification.
\newblock \emph{arXiv preprint arXiv:1511.08630}.

\bibitem[{Ziegele et~al.(2018)Ziegele, Daxenberger, Quiring, and
  Gurevych}]{TUD-CS-2018-0023}
Marc Ziegele, Johannes Daxenberger, Oliver Quiring, and Iryna Gurevych. 2018.
\newblock Developing automated measures to predict incivility in public online
  discussions on the facebook sites of established news media.
\newblock In \emph{Proceedings of the 68th Annual Conference of the
  International Communication Association (ICA)}.

\end{thebibliography}
\bibliographystyle{acl_natbib_nourl}

\end{document}